\documentclass{article}

\def\isdraft{1}

\usepackage{iclr2026_conference,times}
\usepackage[T1]{fontenc}
\usepackage{microtype}
\usepackage{amsmath}
\usepackage{amssymb}
\usepackage{mathtools}
\usepackage{booktabs}
\usepackage{multirow}
\usepackage{graphicx}
\usepackage{placeins}
\usepackage{xcolor}
\usepackage[colorlinks=true,allcolors=blue]{hyperref}
\usepackage[nameinlink,capitalize,noabbrev]{cleveref}

\definecolor{tbdred}{RGB}{180,25,25}

\newcommand{\tbd}[1]{%
  \ifnum\isdraft=1
    \textcolor{tbdred}{\textbf{[TBD: #1]}}%
  \else
    \PackageError{pnpo-paper}{Unresolved placeholder: #1}{Resolve every TBD before release.}%
  \fi
}

\newcommand{\papertitle}{Reusing Rollouts under Policy Lag: Prefix-Normalized Policy Optimization for LLM Reinforcement Learning}

\newcommand{\publicauthors}{%
  \makebox[\dimexpr\textwidth-2\tabcolsep\relax][c]{%
    \begin{tabular}{c}
      Wenhao Zhang$^{1,2}$\enspace
      Yibo Xie$^{1}$\enspace
      Rui Wang$^{1}$\enspace
      Jiahua Yang$^{1,3}$\enspace
      Lei Jiang$^{1,4}$\enspace
      Zibo Yang$^{1}$ \\[0.18em]
      Yawei Wang$^{1}$\enspace
      Jiali Xu$^{1}$\enspace
      jasperawang$^{1}$\enspace
      Haoyang Long$^{1}$\enspace
      Huan Xiong$^{1,\dagger}$\enspace
      alantzhao$^{1,\dagger}$ \\[0.48em]
      \normalfont\small
      $^{1}$Tencent
      \quad $^{2}$Harbin Institute of Technology
      \quad $^{3}$Jinan University \\[-0.02em]
      \normalfont\small
      $^{4}$University of Science and Technology of China \\[-0.02em]
      \normalfont\small
      $^{\dagger}$Corresponding authors
    \end{tabular}%
  }%
}

\newcommand{\publicpdfauthors}{Wenhao Zhang, Yibo Xie, Rui Wang, Jiahua Yang,
Lei Jiang, Zibo Yang, Yawei Wang, Jiali Xu, jasperawang, Haoyang Long,
Huan Xiong, and alantzhao}

\newcommand{\PNPO}{\textsc{PNPO}}
\newcommand{\GSPO}{\textsc{GSPO}}
\newcommand{\GRPO}{\textsc{GRPO}}
\newcommand{\CTPO}{\textsc{CTPO}}
\newcommand{\TEPO}{\textsc{TEPO}}

\newcommand{\pib}{\pi_{\beta}}
\newcommand{\pit}{\pi_{\theta}}
\newcommand{\ratio}{\rho}
\newcommand{\cumratio}{C}
\newcommand{\pnweight}{w^{\mathrm{PN}}}
\newcommand{\adv}{\widehat{A}}
\newcommand{\sg}{\operatorname{sg}}

\title{\papertitle}
\author{\publicauthors}

\iclrfinalcopy

\begin{document}

\maketitle
\fancyhead{}
\hypersetup{
  pdftitle={\papertitle},
  pdfauthor={\publicpdfauthors}
}

\begin{abstract}
Autoregressive rollout generation is a major computational cost in reinforcement
learning for large language models.  Reusing each rollout batch for additional
learner updates amortizes this cost, but later updates become increasingly
off-policy as the learner departs from the behavior policy.  At a token
position, exact off-policy correction must account for both the current action
and the probability of reaching its prefix.  The cumulative importance ratio
provides this correction, but its product form can produce an unwieldy dynamic
range.
We study \emph{Prefix-Normalized Policy Optimization} (\PNPO), which replaces
the cumulative ratio with the geometric mean of likelihood ratios along each
causal prefix, preserving causal-prefix dependence at each position while
compressing the log-weight scale.  In controlled long-context mathematical
reasoning experiments, we induce two off-policy regimes by using one or four
policy-update epochs per rollout batch.  \PNPO{} does not consistently
outperform \GSPO{} with one epoch.  With four epochs, it attains the highest
observed Avg@32 on each benchmark; the unweighted mean of the three
independently selected benchmark peaks is \(50.24\), \(3.00\) percentage
points above \GSPO{}.  Under a matched
\(2{,}400\)-update budget, four-epoch \PNPO{} reaches a final macro Avg@32 of
\(49.66\) after \(150\) rollout batches, comparable to the \(49.56\) reached
after \(600\) batches with one epoch.  These results provide preliminary
evidence that \PNPO{} can be advantageous as training moves further off-policy.
\end{abstract}

\section{Introduction}
\label{sec:introduction}

\begin{figure}[t]
  \centering
  \includegraphics[width=\linewidth]{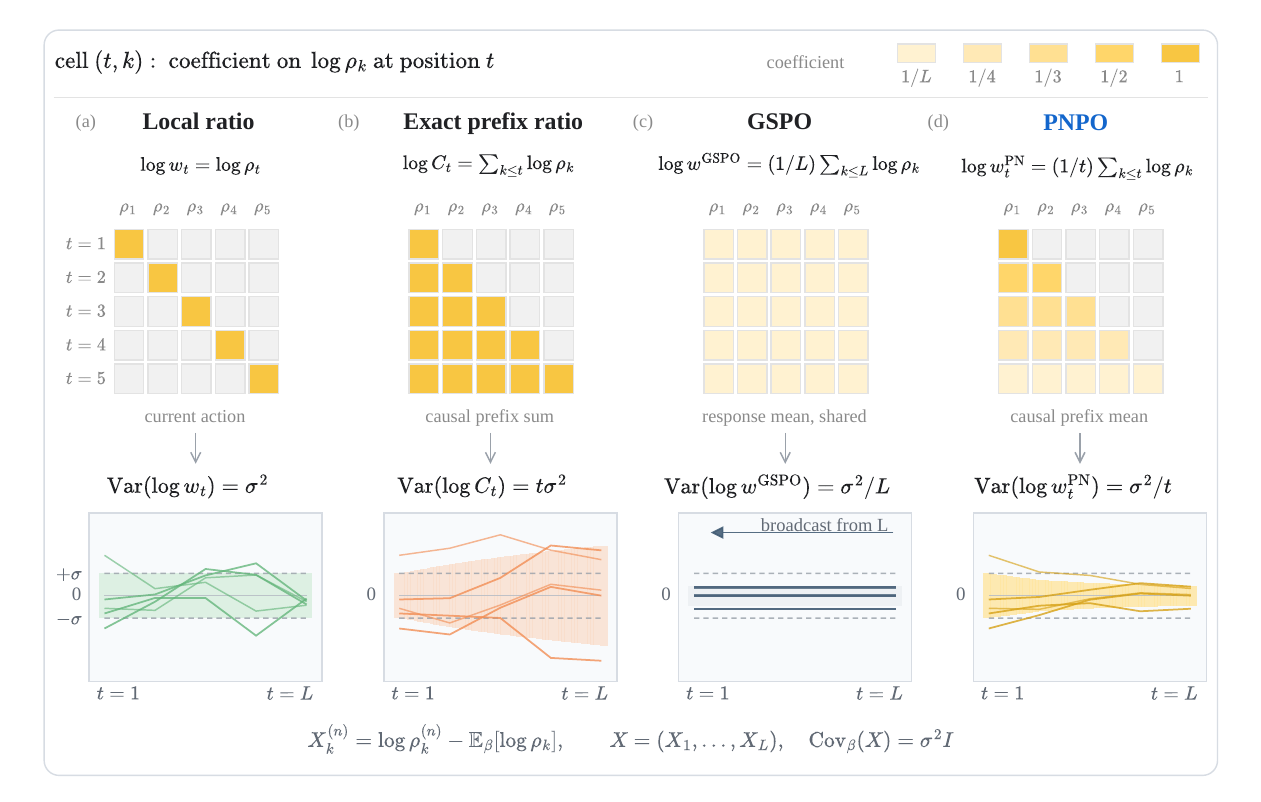}
  \caption{\textbf{Weighting support and log-weight scale.}
  Top: the local log-ratio contains only the current-token term; the exact
  cumulative log-ratio sums terms over the causal prefix; \GSPO{} broadcasts
  the full-response mean log-ratio to every position; and \PNPO{} uses the
  corresponding prefix mean at each position.  Bottom: schematic
  transformations of the same centered log-ratio vectors under the covariance
  model shown in the figure.  Shading shows the corresponding
  \(\pm\sqrt{\operatorname{Var}}\) scale, and dashed lines mark the common
  \(\pm\sigma\) reference.  Among these, the cumulative prefix ratio is the
  exact joint state--action change-of-measure weight at position \(t\); the
  local ratio is exact only for the conditional action distribution at a fixed
  prefix.  \PNPO{} is a scale-controlled, generally biased transform of the
  cumulative ratio.  The traces are schematic, not training curves.}
  \label{fig:weighting-granularities}
\end{figure}

Policy-gradient methods are a standard approach to language-model post-training
\citep{schulman2017ppo,ouyang2022instructgpt,guo2025deepseekr1}.  In this
setting, collecting a rollout batch requires autoregressive generation,
which can consume a substantial fraction of training time
\citep{hu2026dora}.  One way to amortize this cost is to reuse collected
rollouts for multiple learner updates, as in PPO's multi-epoch minibatch
optimization.  Because the behavior data remain fixed while the learner
changes, later updates become increasingly off-policy.

This mismatch already appears within a single pass through a collected batch,
since later minibatches are processed after earlier learner updates, and it
grows when the same trajectories are revisited for additional epochs.  Such
update-induced drift has been noted in recent large-scale reasoning-model
training \citep{deepseek2025v32} and is a form of \emph{forward policy lag}
\citep{honari2026vaco}.  Proximal policy methods address this mismatch through
a local approximation.  The exact performance-difference
identity depends on the learner's state distribution; TRPO freezes this
distribution at the behavior policy in its surrogate and uses a trust region
to control the resulting approximation error \citep{schulman2015trpo}.  PPO
replaces the constrained update with a clipped local-ratio objective
\citep{schulman2017ppo}, and \GRPO{} retains this token-local construction while
replacing critic-based advantages with group-relative estimates
\citep{shao2024deepseekmath}.  The resulting local-ratio objective is therefore
a proximity-based surrogate rather than a complete state--action correction,
and repeated reuse progressively weakens the premise that the learner remains
close to the behavior policy.

Autoregressive generation, however, makes the state correction omitted by this
local surrogate directly computable.  At position \(t\), the state is the full
prefix \(s_t=(x,y_{<t})\), and the deterministic token-appending transition
makes the probability of reaching \(s_t\) factorize over preceding token
decisions.  The local ratio
\(\ratio_t=\pit(y_t\mid x,y_{<t})/\pib(y_t\mid x,y_{<t})\) corrects the action
distribution at the observed prefix, whereas the exact state--action density
ratio is \(\cumratio_t=\prod_{k=1}^{t}\ratio_k\): the preceding factors correct
prefix visitation and \(\ratio_t\) corrects the current action
\citep{precup2000offpolicy,liu2020statecorrection}.  This restores causal-prefix
dependence, but \(\log\cumratio_t\) accumulates token-level log-ratios, so its
dynamic range can grow with prefix length and policy drift.  \GSPO{} instead
controls scale with a length-normalized full-response statistic shared across
all positions \citep{zheng2025gspo}, at the cost of making early-position
weights depend on the sampled suffix.  This leaves a tension between causal
prefix correction and manageable weight scale.

\emph{Prefix-Normalized Policy Optimization} (\PNPO{}) uses the geometric mean
of the likelihood ratios along the prefix through position \(t\):
\begin{equation}
  \pnweight_t
  =
  \cumratio_t^{1/t}
  =
  \exp\left(\frac{1}{t}\sum_{k=1}^{t}\log\ratio_k\right).
  \label{eq:pnpo-ratio-intro}
\end{equation}
The \(1/t\) power tempers the current-token ratio together with all preceding
ratios, trading the exact state--action density ratio for scale
control.
Unlike a sequence-shared statistic, \(\pnweight_t\) varies with position and
excludes likelihood-ratio changes in the future suffix; at the terminal
position, it coincides numerically with the full-response geometric mean.
\Cref{fig:weighting-granularities} summarizes the four weighting granularities
and their position-dependent log-weight scales.  The \PNPO{} objective
evaluated in this paper combines this weight with a position-dependent
acceptance gate and response-level averaging;
\cref{sec:method} gives the full specification.

Our central empirical question is whether \PNPO{} performs more favorably than
\GSPO{} as training moves further off-policy.  We create two
controlled off-policy regimes by applying one or four policy-update epochs to
each rollout batch in long-context mathematical reasoning training.  Within
each regime, we match the training prompts, number of generated responses,
update schedule, and evaluation horizon.  With one epoch, \PNPO{} does not
consistently outperform \GSPO{}.  With four epochs, \PNPO{} attains the highest
observed Avg@32 on each benchmark; the unweighted mean of the three independently
selected benchmark peaks is \(50.24\), \(3.00\) percentage points above
\GSPO{}.  Under the matched \(2{,}400\)-update budget, four-epoch \PNPO{}
reaches a final macro Avg@32 of \(49.66\) after \(150\) rollout batches,
comparable to the \(49.56\) reached after \(600\) batches with one epoch.
Together, these results suggest that \PNPO{} can be particularly useful as
training moves further off-policy. \textbf{Our main contributions are
summarized as follows:}

\noindent\hangindent=1.2em\hangafter=1
\textbullet\hspace{0.5em}We formulate \PNPO{}, a token-level policy objective
for off-policy language-model RL that uses a prefix-normalized policy weight.
Starting from the exact autoregressive state--action ratio, we characterize
this biased, position-specific transformation and compare it with per-token
ratios and sequence-shared weights.
\par
\noindent\hangindent=1.2em\hangafter=1
\textbullet\hspace{0.5em}We evaluate \PNPO{} in two controlled off-policy
regimes induced by one and four PPO epochs, and find that its advantage over
\GSPO{} is more pronounced in the four-epoch regime under a matched
minibatch-update budget.
\par

\section{Problem Setup and Background}
\label{sec:background}

\subsection{Local Proximal Surrogate}

Fix a prompt \(x\) and let the behavior policy \(\pib\) generate a response
\(y_{1:L}\).  At position \(t\), the state is the complete prefix
\(s_t=(x,y_{<t})\), the action is \(a_t=y_t\), and \(d_t^\beta\) and
\(d_t^\theta\) denote the state distributions induced by the behavior and
learner policies, respectively.  Let \(A_t^\beta\) be the finite-horizon
advantage under \(\pib\), and define the local action ratio
\(\ratio_t=\pit(a_t\mid s_t)/\pib(a_t\mid s_t)\).

The advantage \(A_t^\beta\) enters through the performance-difference identity,
which expresses the change from \(\pib\) to \(\pit\) using behavior-policy
advantages but learner-policy state--action visitation.  TRPO obtains a
tractable local objective by freezing the learner state distribution at
\(d_t^\beta\) and using \(\ratio_t\) to change only the conditional action
distribution.  The corresponding unclipped surrogate gradient is
\begin{equation}
  \nabla_\theta\mathcal{S}_{\beta,x}(\theta)
  =
  \sum_{t=1}^{L}
  \mathbb{E}_{\substack{s_t\sim d_t^\beta(\cdot\mid x)\\
                        a_t\sim\pib(\cdot\mid s_t)}}
  \left[
    \ratio_t A_t^\beta(s_t,a_t)
    \nabla_\theta\log\pit(a_t\mid s_t)
  \right].
  \label{eq:local-surrogate-gradient}
\end{equation}
Here, \(A_t^\beta\) and \(d_t^\beta\) are fixed, and differentiating the local
ratio yields the current-token score.  The ratio \(\ratio_t\) therefore changes
the action distribution exactly at a fixed state, but it does not change state
weighting from \(d_t^\beta\) to \(d_t^\theta\).  This omission defines
a proximal approximation rather than an exact state--action correction.  The
displayed gradient agrees with the true policy gradient at \(\pit=\pib\); away
from the behavior policy, the accuracy of the surrogate depends on controlling
policy movement.

TRPO controls this frozen-occupancy approximation with a trust region
\citep{schulman2015trpo}.  PPO replaces the explicit constrained update with a
clipped local-ratio surrogate \citep{schulman2017ppo}, and \GRPO{} retains the
same token-level ratio and clipping while replacing critic-based advantages with
a group-relative outcome signal \citep{shao2024deepseekmath}.  These methods
rely on policy proximity to limit the error induced by state-distribution shift
rather than explicitly applying the missing state ratio.  Reusing a rollout
batch for additional learner updates can progressively weaken this proximity
condition.  The performance-difference construction and the resulting local
surrogate are derived in \cref{app:derivation}.

\subsection{Exact Prefix Change of Measure}

In a general MDP, the state-distribution ratio omitted by the local surrogate is
difficult to compute.  In the autoregressive process considered here, however,
the full prefix is observed and the transition deterministically appends the
sampled token.  The probability of reaching \(s_t\) therefore factorizes along
its unique prefix path.  Under the standard support condition,
\begin{align}
  \frac{d_t^\theta(s_t\mid x)}
       {d_t^\beta(s_t\mid x)}
  &=
  \prod_{k=1}^{t-1}\ratio_k,
  \nonumber\\[1mm]
  \frac{d_t^\theta(s_t\mid x)\pit(a_t\mid s_t)}
       {d_t^\beta(s_t\mid x)\pib(a_t\mid s_t)}
  &=
  \underbrace{\prod_{k=1}^{t-1}\ratio_k}_{\text{state ratio}}
  \underbrace{\ratio_t}_{\text{action ratio}}
  \nonumber\\
  &=
  \prod_{k=1}^{t}\ratio_k
  \eqqcolon
  \cumratio_t.
  \label{eq:state-action-factorization}
\end{align}
The factors before \(t\) correct the probability of reaching the current
prefix, while \(\ratio_t\) corrects the action selected there.  Thus,
\(\cumratio_t\) is the exact joint state--action change-of-measure ratio at
position \(t\).  Under the same support condition, for any integrable function
\(f_t(s_t,a_t)\),
\begin{equation}
  \mathbb{E}_{\substack{s_t\sim d_t^\beta(\cdot\mid x)\\
                        a_t\sim\pib(\cdot\mid s_t)}}
  \left[\cumratio_t f_t(s_t,a_t)\right]
  =
  \mathbb{E}_{\substack{s_t\sim d_t^\theta(\cdot\mid x)\\
                        a_t\sim\pit(\cdot\mid s_t)}}
  \left[f_t(s_t,a_t)\right].
  \label{eq:prefix-change-of-measure}
\end{equation}
This is the autoregressive form of per-decision importance sampling
\citep{precup2000offpolicy,zhang2026ctpo}; its state-ratio factor is the
state-distribution correction required for exact off-policy gradient estimation
\citep{liu2020statecorrection}.  The factorization and generic change-of-measure
identity are derived in \cref{app:derivation}.

To recover the target-policy gradient rather than the frozen-occupancy
surrogate, let \(J_x(\pi)\) denote the expected return for prompt \(x\), let
\(A_t^\theta\) be the finite-horizon advantage under \(\pit\), and define
\(z_t=\nabla_\theta\log\pit(a_t\mid s_t)\).  Applying
\cref{eq:prefix-change-of-measure} to the policy-gradient theorem
\citep{sutton1999policygradient} gives
\begin{align}
  \nabla_\theta J_x(\pit)
  &=
  \sum_{t=1}^{L}
  \mathbb{E}_{\substack{s_t\sim d_t^\theta(\cdot\mid x)\\
                        a_t\sim\pit(\cdot\mid s_t)}}
  \left[A_t^\theta(s_t,a_t)z_t\right]
  \nonumber\\
  &=
  \sum_{t=1}^{L}
  \mathbb{E}_{\substack{s_t\sim d_t^\beta(\cdot\mid x)\\
                        a_t\sim\pib(\cdot\mid s_t)}}
  \left[\cumratio_t A_t^\theta(s_t,a_t)z_t\right].
  \label{eq:current-token-gradient-change-of-measure}
\end{align}

The exact cumulative ratio can also create a scale problem.  Because
\(\log\cumratio_t=\sum_{k=1}^{t}\log\ratio_k\), its dynamic range can grow with
prefix length and policy drift.  Token-level PPO and \GRPO{} avoid this
accumulation by using only \(\ratio_t\), but thereby omit the prefix-state
factor.  \GSPO{} instead uses a length-normalized full-response statistic shared
across positions, controlling response-level scale while making early-token
weights depend on the sampled suffix \citep{zheng2025gspo}.  \PNPO{} uses the
position-dependent transform \(\cumratio_t^{1/t}\), retaining causal-prefix
dependence while compressing cumulative log scale.  For \(t>1\), this transform
is not the target-to-behavior density ratio and therefore does not preserve the
exact change of measure in \cref{eq:prefix-change-of-measure}; it trades
exactness for scale control.  Exact change of measure alone, however,
does not determine a compatible advantage--score pairing.

\subsection{Exact Advantage--Score Pairings}

Let \(\mathbb{E}_{\pib}[\cdot]\) denote expectation over complete responses
sampled from \(\pib(\cdot\mid x)\).  Equation
\eqref{eq:current-token-gradient-change-of-measure} is an exact target-policy
gradient representation and pairs the current-token score \(z_t\) with
\(A_t^\theta\).  An unbiased representation that retains \(A_t^\beta\) must
instead be derived from an exact identity, rather than obtained by substituting
\(A_t^\beta\) for \(A_t^\theta\) in that equation.  The familiar
behavior-advantage/current-token-score pairing in PPO is the gradient structure
of the frozen-occupancy surrogate in
\cref{eq:local-surrogate-gradient}.  If \(\cumratio_t\) is substituted only as
a detached coefficient while retaining this current-token-score structure, the
resulting expression is
\[
  \sum_{t=1}^{L}
  \mathbb{E}_{\pib}
  \left[\cumratio_t A_t^\beta(s_t,a_t)z_t\right],
\]
which is generally not equal to \(\nabla_\theta J_x(\pit)\).  This retains the
gradient structure of the PPO surrogate rather than differentiating an exact
identity.  To retain \(A_t^\beta\) without bias, we instead differentiate the
exact behavior-rollout performance-difference identity, obtaining
\begin{equation}
  \nabla_\theta J_x(\pit)
  =
  \mathbb{E}_{\pib}
  \left[
    \sum_{t=1}^{L}
    \cumratio_t A_t^\beta(s_t,a_t)
    \sum_{k=1}^{t}z_k
  \right].
  \label{eq:exact-gradient-pairings}
\end{equation}
Thus, retaining \(A_t^\beta\) requires the cumulative prefix score
\(\sum_{k=1}^{t}z_k\), whereas the current-token score in
\cref{eq:current-token-gradient-change-of-measure} pairs with
\(A_t^\theta\).  Under the support condition and with exact advantages, these
representations are unbiased and equal in expectation, but their components
are not interchangeable.  \Cref{app:gradient-representations} proves this
equivalence.

The practical \PNPO{} objective retains the current-token-score structure of
PPO and \GRPO{}, using the rollout-derived group-relative advantage as a
behavior-side proxy rather than the exact \(A_t^\theta\).  It also replaces the
exact cumulative ratio with \(\cumratio_t^{1/t}\) and combines this weight with
position-dependent gating and response-level aggregation.  The complete
\PNPO{} objective is therefore a biased proximal surrogate, while
\(\cumratio_t\) remains its exact change-of-measure reference.
\Cref{sec:method} gives the full objective; the derivation and the boundary
between the exact and approximate forms are detailed in
\cref{app:gradient-representations}.

\section{Prefix-Normalized Policy Optimization}
\label{sec:method}

\subsection{Prefix-Normalized Policy Weight}

Starting from the exact state--action ratio in
\cref{eq:state-action-factorization}, \PNPO{} length-normalizes the cumulative
log-ratio by the number of decisions in the corresponding prefix.  For a
response \(y_i=y_{i,1:L_i}\) with valid length \(L_i\), at position \(t\) we
define
\begin{equation}
  \pnweight_{i,t}
  =
  \cumratio_{i,t}^{1/t}
  =
  \exp\left(\frac{1}{t}\sum_{k=1}^{t}\log\ratio_{i,k}\right).
  \label{eq:pnpo-ratio}
\end{equation}
The \(1/t\) power applies to the current-token ratio together with every
preceding ratio.  For \(t>1\), \(\pnweight_{i,t}\) is not a
target-to-behavior density ratio and therefore does not preserve the exact
change of measure associated with \cref{eq:state-action-factorization}.  At
each fixed \(t\), it is a
monotone transform of \(\cumratio_{i,t}\): it preserves the sign of
\(\log\cumratio_{i,t}\) and the ordering across responses while compressing the
cumulative log scale.  Given the prompt \(x_i\), the weight depends only on
\(y_{i,\leq t}\).

The weight recovers the local ratio at \(t=1\) and the full-response geometric
mean at \(t=L_i\), while intermediate positions exclude likelihood shifts from
the future suffix.  We refer to \(\pnweight_{i,t}\) as the
\emph{prefix-normalized policy weight}.

\subsection{Complete PNPO Objective}
\label{sec:pnpo-objective}

\paragraph{Acceptance gate.}
The reported \PNPO{} configuration uses the position-dependent scale
\begin{equation}
  h(t,L_i)=\sqrt{\frac{L_i}{t}},
  \label{eq:position-function}
\end{equation}
and retains the score term at position \(t\) only when
\begin{equation}
  M_{i,t}
  =
  \mathbf{1}\left\{
  1-\epsilon_-\,h(t,L_i)
  \leq \pnweight_{i,t}
  \leq 1+\epsilon_+\,h(t,L_i)
  \right\}.
  \label{eq:position-gate}
\end{equation}
This heuristic widens the acceptance interval at earlier positions and reduces
to the base interval at \(t=L_i\); the lower and upper base tolerances may be
asymmetric.  The bounds are evaluated independently at each position.  If
\(M_{i,t}=0\), only the score-function term at position \(t\) is removed, and
later positions remain eligible under their own bounds.  The gate performs hard
rejection rather than truncating the weight to a boundary.

\paragraph{Group-relative advantage and detached surrogate.}
Let \(x\sim\mathcal{D}\) be a prompt, and sample \(G\) conditionally
independent responses \(y_i\sim\pib(\cdot\mid x)\).  Writing
\(R_i=R(x,y_i)\), the reported experiments use \GRPO{}'s group-relative
outcome advantage \citep{shao2024deepseekmath}, assigning every valid token in
response \(i\) the advantage
\begin{equation}
  \adv_{i,t}
  =
  \frac{
    R_i-\operatorname{mean}\!\left(\{R_j\}_{j=1}^{G}\right)
  }{
    \operatorname{std}\!\left(\{R_j\}_{j=1}^{G}\right)
  }.
  \label{eq:group-relative-advantage}
\end{equation}
We maximize the detached score-function surrogate
\begin{equation}
  \mathcal{J}_{\PNPO}(\theta)
  =
  \mathbb{E}_{\substack{x\sim\mathcal{D}\\
  y_1,\ldots,y_G\overset{\mathrm{i.i.d.}}{\sim}\pib(\cdot\mid x)}}
  \left[
    \frac{1}{G}\sum_{i=1}^{G}\frac{1}{L_i}
    \sum_{t=1}^{L_i}
    \sg\left[M_{i,t}\pnweight_{i,t}\adv_{i,t}\right]
    \log\pit(y_{i,t}\mid x,y_{i,<t})
  \right].
  \label{eq:pnpo-objective}
\end{equation}
The stop-gradient operator treats the gate, policy weight, and advantage as
numerical coefficients, so gradients flow only through the current-token log
probability.  We first average the token-level score terms over valid tokens
within each response and then average the resulting per-response means across
responses, matching the aggregation induced by the \GSPO{} objective
\citep{zheng2025gspo}.  The normalization remains \(1/L_i\) after gating:
rejected tokens contribute zero, and the retained tokens are not renormalized
by their accepted count.  Only \(\pnweight_{i,t}\) depends exclusively on the
observed prefix; \(M_{i,t}\) depends on \(L_i\), and \(\adv_{i,t}\) is derived
from response-level outcomes.

\paragraph{Updates over a fixed rollout batch.}
The sampled responses, behavior-policy log probabilities in the denominator,
and group-derived advantages remain fixed while a rollout batch is reused.  At
each minibatch update, including those in later epochs, we evaluate the
numerator log probabilities under the current learner, reconstruct
\(\pnweight_{i,t}\) from the cumulative log-ratios, recompute \(M_{i,t}\), and
apply \cref{eq:pnpo-objective}.  In each additional epoch, the learner is
updated further on the same sampled responses without generating new
trajectories.

\section{Experiments}
\label{sec:experiments}

We evaluate \PNPO{} on long-context mathematical reasoning using one or four
PPO epochs per rollout batch.  We compare primarily against \GSPO{} and include
\GRPO{} as a token-level baseline.

\subsection{Experimental Setup}

\paragraph{Datasets and model.}
We initialize all methods from DeepSeek-R1-Distill-Qwen-1.5B
\citep{guo2025deepseekr1} and perform RL post-training on DAPO-Math-17k
\citep{yu2025dapo}.  We evaluate mathematical reasoning on AMC 2023,
AIME 2024, and AIME 2025.

\paragraph{Training setup.}
At each training step, we collect one rollout batch by sampling \(256\) prompts
and \(8\) responses per prompt, yielding \(2{,}048\) responses.  The maximum
prompt and response lengths are \(1{,}024\) and \(15{,}360\) tokens,
respectively.  We use a learning rate of \(10^{-6}\) and minibatches of \(64\)
prompt groups (\(512\) responses), yielding four optimizer updates per PPO
epoch.  We run \(600\) steps with one epoch and \(150\) steps with four epochs,
so both settings perform \(2{,}400\) optimizer updates.  All training runs use
\(32\) NVIDIA H20 GPUs.  Further implementation details are provided in
\cref{app:configuration}.

\paragraph{Baselines.}
\GRPO{} uses local token ratios with token-mean aggregation
\citep{shao2024deepseekmath}, lower/upper clipping tolerances
\((0.2,0.28)\), and a dual-clip coefficient of \(10\).  \GSPO{} uses a
length-normalized response ratio with response-mean aggregation
\citep{zheng2025gspo} and lower/upper clipping tolerances
\((3\times10^{-4},4\times10^{-4})\).  \PNPO{} uses the complete objective in
\cref{sec:pnpo-objective}, with base acceptance-gate tolerances
\((7\times10^{-4},9.5\times10^{-4})\).

\paragraph{Evaluation.}
For each benchmark, Avg@32 is the mean binary correctness over \(32\) sampled
responses per problem; macro Avg@32 is the unweighted mean of the three
benchmark scores.  We evaluate every \(50\) steps in the one-epoch runs and
every \(10\) steps in the four-epoch runs.  For each benchmark,
\cref{tab:main-results} reports the best observed Avg@32 within the
corresponding training horizon.

\subsection{Main Results}

Under four PPO epochs, \PNPO{} attains the best observed Avg@32 on all three
benchmarks (\cref{tab:main-results}).  The unweighted average of these three
scores is \(50.24\), \(3.00\) percentage points above \GSPO{}.  We next examine
whether this separation persists over the reported evaluation trajectories.

\begin{table}[t]
  \centering
  \caption{\textbf{Best observed Avg@32 (\%) within each training horizon.}
  For each run, the best value on each benchmark is selected independently;
  Avg.\ is the unweighted mean of the three resulting scores.  Boldface marks
  the best value in each column within an epoch setting.}
  \label{tab:main-results}
  \renewcommand{\arraystretch}{1.08}
  \setlength{\tabcolsep}{4pt}
  \begin{tabular*}{\linewidth}{@{\extracolsep{\fill}}clcccc@{}}
    \toprule
    \textbf{PPO Epochs} & \textbf{Method} & \textbf{AMC23} &
    \textbf{AIME24} & \textbf{AIME25} & \textbf{Avg.} \\
    \midrule
    \multirow{3}{*}{1}
      & \GRPO & \textbf{80.16} & 35.83 & 28.75 & 48.25 \\
      & \GSPO & 79.92 & 37.81 & \textbf{30.52} & 49.42 \\
      & \PNPO{} (Ours) & 79.84 & \textbf{40.52} & 29.79 &
        \textbf{50.05} \\
    \midrule
    \multirow{3}{*}{4}
      & \GRPO & 77.81 & 36.15 & 27.19 & 47.05 \\
      & \GSPO & 78.91 & 34.90 & 27.92 & 47.24 \\
      & \PNPO{} (Ours) & \textbf{80.94} & \textbf{38.85} &
        \textbf{30.94} & \textbf{50.24} \\
    \bottomrule
  \end{tabular*}
\end{table}

\begin{figure}[t]
  \centering
  \includegraphics[width=\linewidth]{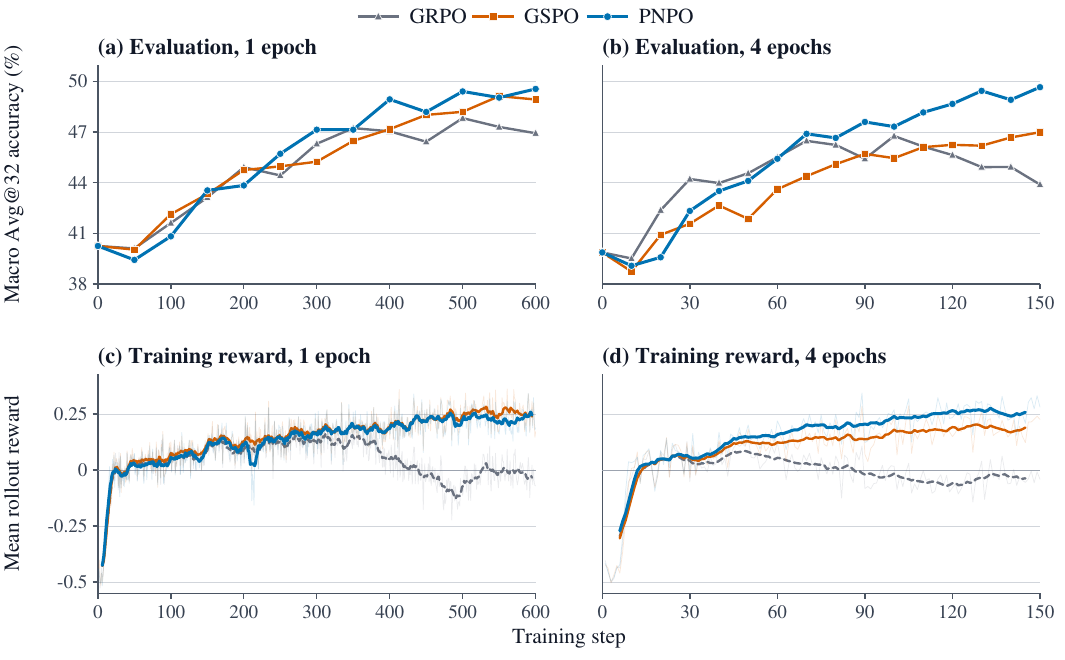}
  \caption{\textbf{Evaluation and training dynamics under one and four PPO
  epochs.}  (a,b) Macro Avg@32 (\%), computed as the unweighted mean across
  AMC 2023, AIME 2024, and AIME 2025, without smoothing.  (c,d) Per-step mean
  rollout reward under the corresponding settings; faint curves show raw
  rewards and darker curves show centered 11-step moving averages.  All runs
  begin from the same checkpoint, whose \GRPO{} evaluation is used as the
  common step-\(0\) value.}
  \label{fig:experiment-curves}
\end{figure}

\subsection{Learning Dynamics and Rollout Reuse}

\Cref{fig:experiment-curves} shows that the separation in
\cref{tab:main-results} is regime-dependent.  With one epoch, \PNPO{} and
\GSPO{} remain close and exchange the lead.  With four epochs, \PNPO{} is higher
at \(14\) of \(15\) evaluations and finishes \(2.66\) percentage points ahead of
\GSPO{}.  The four-epoch advantage is therefore visible across the reported
evaluation trajectory, rather than only in the per-benchmark peaks.

At the common budget of \(2{,}400\) optimizer updates, four-epoch \PNPO{}
reaches a final macro Avg@32 of \(49.66\) after \(150\) rollout batches,
comparable to the \(49.56\) reached after \(600\) one-epoch batches.  Because
the rollout batch size is fixed, the four-epoch run reaches comparable final
performance using one quarter as many newly generated responses, indicating
more effective rollout reuse in this setting.

\begin{figure}[t]
  \centering
  \includegraphics[width=\linewidth]{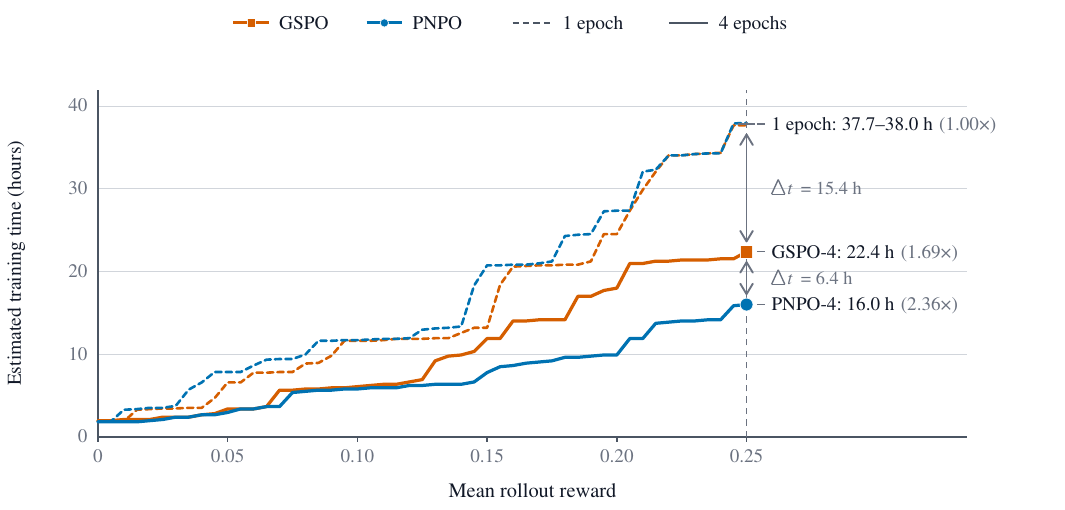}
  \caption{\textbf{Estimated time to reach a mean rollout reward of \(0.25\).}
  First-passage times are computed from the centered 11-step reward
  trajectories using \(283\) and \(510\) seconds per step in the one- and
  four-epoch settings, respectively.}
  \label{fig:time-to-reward}
\end{figure}

Under the regime-level timing estimates used in \cref{fig:time-to-reward}, the
centered reward trajectories first reach \(0.25\) after \(37.7\)--\(38.0\) hours
in the one-epoch runs, \(22.4\) hours for four-epoch \GSPO{}, and \(16.0\) hours
for four-epoch \PNPO{}.  Both four-epoch configurations therefore reach the
threshold earlier than their one-epoch counterparts, while \PNPO{} reaches it
\(6.4\) hours earlier than \GSPO{} with four epochs.

\GRPO{} does not sustain its early reward gains over either training horizon,
and its four-epoch evaluation curve declines after reaching its maximum.
\GSPO{} provides a useful counterpoint: its four-epoch reward trajectory
remains stable, but this stability does not eliminate the evaluation gap.  This
pattern is consistent with the weighting-granularity interpretation: at each
update, \GSPO{} shares one response-level statistic across token positions,
whereas \PNPO{} retains a position-dependent prefix statistic.  Because the
experiments compare the complete objectives without an ablation of weighting
granularity, however, they do not isolate this mechanism.

\FloatBarrier
\section{Related Work}
\label{sec:related}

\paragraph{Off-policy language-model reinforcement learning.}
Learner--behavior mismatch arises whenever a policy is updated using
trajectories generated by an earlier policy.  PPO-style minibatch updates can
make even freshly collected rollouts off-policy, while training--inference
inconsistencies introduce a related mismatch
\citep{schulman2017ppo,deepseek2025v32}.  Replay buffers resample
older trajectories across learner updates \citep{arnal2026replay}, while OAPL
introduces an explicitly off-policy objective for data from a lagged inference
policy \citep{ritter2026oapl}.  Concurrent generation and optimization
introduce additional policy-version staleness in asynchronous actor--learner
systems \citep{espeholt2018impala,fu2025areal,hu2026dora}; VACO and SAO address
policy lag through algorithmic modifications
\citep{honari2026vaco,hou2026sao}.  These settings share an
importance-weighting problem, although the mismatch arises in different ways.
Our experiments isolate one controlled source of mismatch by varying
the number of learner updates applied to each fixed rollout batch; replay and
asynchronous staleness remain outside the scope of this study.

\paragraph{Likelihood-ratio granularity in reasoning RL.}
Recent critic-free reasoning objectives differ in advantage estimation,
clipping, and likelihood-ratio granularity.  \GRPO{} and DAPO use per-token
ratios with group-relative outcome advantages
\citep{shao2024deepseekmath,yu2025dapo}, whereas \GSPO{} uses a
length-normalized response ratio in a sequence-level objective
\citep{zheng2025gspo}.  \PNPO{} occupies the prefix-dependent middle ground
between these token- and response-level constructions.

\paragraph{Prefix-aware importance weighting.}
Classical per-decision importance sampling provides the foundation for
prefix-conditioned correction \citep{precup2000offpolicy}.  Recent LLM methods
construct prefix-aware importance weights in different ways.  MinPRO replaces
the cumulative product over preceding tokens with the minimum of their
likelihood ratios while retaining the current-token ratio
\citep{lei2026minpro}.  \CTPO{} uses the exact cumulative prefix ratio with
position-adaptive clipping \citep{zhang2026ctpo}, whereas \TEPO{} evaluates a
length-normalized prefix likelihood-ratio variant as its ``Sentence Prefix IS''
ablation \citep{lin2026tepo}.  In \PNPO{}, the prefix geometric mean is the
primary policy weight, normalizing the cumulative log-ratio by prefix length
rather than preserving exact change of measure.

\section{Conclusion and Limitations}
\label{sec:conclusion}

We introduced \PNPO{}, a token-level objective that replaces the exact
cumulative state--action ratio with the geometric mean of token likelihood
ratios over the causal prefix.  This biased normalization
preserves causal prefix dependence at each position while compressing the
cumulative log-ratio scale.  In our controlled comparison of one and four PPO
epochs, \PNPO{} and \GSPO{} perform similarly in the lower-mismatch setting,
whereas \PNPO{} shows a sustained advantage when repeated updates induce greater
learner--behavior mismatch.  At a matched budget of \(2{,}400\) optimizer
updates, four-epoch \PNPO{} reaches final performance comparable to its
one-epoch counterpart while using one quarter as many newly generated
responses, providing preliminary evidence of more effective rollout reuse
under greater mismatch.

These findings are limited to a single \(1.5\)B-parameter model, three
mathematical reasoning benchmarks, and one run per configuration.  Because the
acceptance gate and response-level aggregation are not ablated, the evidence
applies to the complete \PNPO{} configuration and does not isolate the effect
of prefix normalization.  Moreover, our experiments induce mismatch only
through repeated updates on fresh batches.  Asynchronous collection, replay,
offline data, and training--inference mismatch introduce distinct
learner--behavior gaps.

These broader sources of mismatch provide natural settings for future work.
In classical distributed RL, asynchronous actor--learner systems gain parallel
exploration and higher throughput at the cost of policy-lagged experience
\citep{espeholt2018impala}.  Evaluating \PNPO{} in such settings would test
whether prefix-normalized weighting remains useful beyond repeated update
epochs and, in online systems, how more effective reuse of lagged trajectories
interacts with the exploration--exploitation trade-off.

\bibliographystyle{iclr2026_conference}
\bibliography{references}

\clearpage
\appendix
\section{Finite-Horizon Local Surrogates and Exact Prefix Change of Measure}
\label{app:derivation}

Fix a prompt \(x\) and consider an undiscounted autoregressive process with
maximum horizon \(L\).  At position \(t\), let
\(s_t=(x,y_{<t})\) and \(a_t=y_t\).  For a policy \(\pi\), define
\begin{equation}
  d_t^\pi(s\mid x)
  =
  \Pr(s_t=s\mid x,\pi),
  \label{eq:appendix-state-distribution}
\end{equation}
and let \(J_x(\pi)\) and \(A_t^\pi\) denote its expected return and
finite-horizon advantage, respectively.  For responses that terminate before
\(L\), the same derivation applies through the realized terminal position.  We
assume throughout that the learner state--action distribution is absolutely
continuous with respect to the behavior distribution.

\subsection{Local-Ratio Surrogate}

Writing the finite-horizon occupancy position by position, the
performance-difference identity is
\begin{equation}
  J_x(\pit)-J_x(\pib)
  =
  \sum_{t=1}^{L}
  \mathbb{E}_{\substack{s_t\sim d_t^\theta(\cdot\mid x)\\
                        a_t\sim\pit(\cdot\mid s_t)}}
  \left[A_t^\beta(s_t,a_t)\right].
  \label{eq:appendix-performance-difference}
\end{equation}
Define the local action ratio
\begin{equation}
  \ratio_t
  =
  \frac{\pit(a_t\mid s_t)}
       {\pib(a_t\mid s_t)}.
  \label{eq:appendix-local-ratio}
\end{equation}
For a fixed state \(s_t\), the action distribution admits the exact change of
measure
\begin{align}
  &\mathbb{E}_{a_t\sim\pit(\cdot\mid s_t)}
  \left[A_t^\beta(s_t,a_t)\right]
  \nonumber\\
  &\quad =
  \sum_a \pit(a\mid s_t)A_t^\beta(s_t,a)
  \nonumber\\
  &\quad =
  \sum_a \pib(a\mid s_t)
  \frac{\pit(a\mid s_t)}{\pib(a\mid s_t)}
  A_t^\beta(s_t,a)
  \nonumber\\
  &\quad =
  \mathbb{E}_{a_t\sim\pib(\cdot\mid s_t)}
  \left[\ratio_t A_t^\beta(s_t,a_t)\right].
  \label{eq:appendix-action-change}
\end{align}
Proximal policy methods use a local surrogate that freezes each position-wise
state distribution in \cref{eq:appendix-performance-difference} at its
behavior-policy counterpart:
\begin{equation}
  \mathcal{S}_{\beta,x}(\theta)
  =
  J_x(\pib)
  +
  \sum_{t=1}^{L}
  \mathbb{E}_{\substack{s_t\sim d_t^\beta(\cdot\mid x)\\
                        a_t\sim\pib(\cdot\mid s_t)}}
  \left[\ratio_t A_t^\beta(s_t,a_t)\right].
  \label{eq:appendix-local-surrogate}
\end{equation}
Combining \cref{eq:appendix-performance-difference,eq:appendix-action-change}
gives the exact discrepancy
\begin{align}
  J_x(\pit)-\mathcal{S}_{\beta,x}(\theta)
  &=
  \sum_{t=1}^{L}\sum_s
  \left[d_t^\theta(s\mid x)-d_t^\beta(s\mid x)\right]
  \nonumber\\
  &\quad\cdot
  \sum_a \pit(a\mid s)A_t^\beta(s,a).
  \label{eq:appendix-surrogate-gap}
\end{align}
Thus, \(\ratio_t\) exactly changes the conditional action distribution at a
given state, while the surrogate continues to weight states according to
\(d_t^\beta\), rather than \(d_t^\theta\).

\subsection{TRPO Trust Regions and PPO Clipping}

The occupancy discrepancy in \cref{eq:appendix-surrogate-gap} motivates
restricting each policy update.  TRPO optimizes the frozen-occupancy surrogate
subject to a policy-distance constraint; its practical update uses the average
KL constraint
\begin{align}
  \max_\theta\quad
  &\mathcal{S}_{\beta,x}(\theta)
  \nonumber\\
  \text{s.t.}\quad
  &\overline{D}_{\mathrm{KL}}^\beta(\theta)
  \leq \delta,
  \label{eq:appendix-trpo-constraint}
\end{align}
where
\begin{equation}
  \overline{D}_{\mathrm{KL}}^\beta(\theta)
  =
  \frac{1}{L}\sum_{t=1}^{L}
  \mathbb{E}_{s_t\sim d_t^\beta(\cdot\mid x)}
  \left[
    D_{\mathrm{KL}}
    \bigl(
      \pib(\cdot\mid s_t)
      \Vert
      \pit(\cdot\mid s_t)
    \bigr)
  \right].
  \label{eq:appendix-average-kl}
\end{equation}
and \(\delta\) is the trust-region radius.  PPO replaces the explicit
constrained optimization with a clipped local-ratio surrogate
\citep{schulman2017ppo}:
\begin{align}
  \mathcal{J}_{\mathrm{clip}}(\theta)
  =
  \sum_{t=1}^{L}
  \mathbb{E}_{\substack{s_t\sim d_t^\beta(\cdot\mid x)\\
                        a_t\sim\pib(\cdot\mid s_t)}}
  \Big[
  \min\big(&\ratio_t A_t^\beta(s_t,a_t),
  \nonumber\\[-1mm]
  &\operatorname{clip}(\ratio_t,1-\epsilon,1+\epsilon)
  A_t^\beta(s_t,a_t)\big)
  \Big].
  \label{eq:appendix-ppo-clip}
\end{align}
Here, \(\epsilon\) is the clipping tolerance.  Clipping suppresses
advantage-improving updates once a sampled ratio moves beyond the prescribed
interval; it does not impose a hard constraint on the full-policy KL.
\GRPO{} retains the PPO-style token-level action ratio and clipping while
replacing the critic-based advantage with a group-relative outcome signal.

\subsection{Exact Autoregressive State--Action Change of Measure}

The local surrogate in \cref{eq:appendix-local-surrogate} does not explicitly
correct the state-distribution mismatch.  In the autoregressive process
considered here, \(s_t\) contains the complete generated prefix and the
transition deterministically appends the sampled token.  The probability of
reaching \(s_t\) therefore factorizes as
\begin{equation}
  d_t^\pi(s_t\mid x)
  =
  \Pr_\pi(y_{<t}\mid x)
  =
  \prod_{k=1}^{t-1}\pi(a_k\mid s_k).
  \label{eq:appendix-state-factorization}
\end{equation}
Along a sampled response,
\begin{align}
  \frac{d_t^\theta(s_t\mid x)}
       {d_t^\beta(s_t\mid x)}
  &=
  \frac{\prod_{k=1}^{t-1}\pit(a_k\mid s_k)}
       {\prod_{k=1}^{t-1}\pib(a_k\mid s_k)}
  \nonumber\\
  &=
  \prod_{k=1}^{t-1}\ratio_k.
  \label{eq:appendix-state-ratio}
\end{align}
Multiplying by the action ratio at position \(t\) gives the complete
state--action density ratio:
\begin{align}
  &\frac{d_t^\theta(s_t\mid x)\pit(a_t\mid s_t)}
        {d_t^\beta(s_t\mid x)\pib(a_t\mid s_t)}
  \nonumber\\
  &\quad =
  \underbrace{\prod_{k=1}^{t-1}\ratio_k}_{\text{state ratio}}
  \underbrace{\ratio_t}_{\text{action ratio}}
  \nonumber\\
  &\quad =
  \prod_{k=1}^{t}\ratio_k
  \eqqcolon
  \cumratio_t.
  \label{eq:appendix-state-action-ratio}
\end{align}
Consequently, for any integrable function \(f_t(s_t,a_t)\),
\begin{align}
  &\mathbb{E}_{\substack{s_t\sim d_t^\theta(\cdot\mid x)\\
                          a_t\sim\pit(\cdot\mid s_t)}}
  \left[f_t(s_t,a_t)\right]
  \nonumber\\
  &\quad =
  \sum_{s_t}d_t^\theta(s_t\mid x)
  \sum_{a_t}\pit(a_t\mid s_t)f_t(s_t,a_t)
  \nonumber\\
  &\quad =
  \sum_{s_t}d_t^\beta(s_t\mid x)
  \sum_{a_t}\pib(a_t\mid s_t)
  \frac{d_t^\theta(s_t\mid x)\pit(a_t\mid s_t)}
       {d_t^\beta(s_t\mid x)\pib(a_t\mid s_t)}
  f_t(s_t,a_t)
  \nonumber\\
  &\quad =
  \mathbb{E}_{\substack{s_t\sim d_t^\beta(\cdot\mid x)\\
                          a_t\sim\pib(\cdot\mid s_t)}}
  \left[\cumratio_t f_t(s_t,a_t)\right].
  \label{eq:appendix-change-of-measure}
\end{align}
Thus, \(\cumratio_t\) is the exact state--action density ratio at position
\(t\).  \Cref{app:gradient-representations} uses this identity to compare two
exact behavior-rollout representations of the target-policy gradient with the
practical \PNPO{} surrogate.

\section{Exact Gradient Representations and the Practical PNPO Surrogate}
\label{app:gradient-representations}

We retain the setting of \cref{app:derivation}.  The behavior policy \(\pib\)
is fixed, rewards and autoregressive transitions have no explicit dependence
on \(\theta\), and the support condition above holds.  Define the score at
position \(t\) as
\begin{equation}
  z_t
  =
  \nabla_\theta\log\pit(a_t\mid s_t).
  \label{eq:appendix-score}
\end{equation}
For compactness, let
\(\mathbb{E}_{\pib}[\cdot]\coloneqq
\mathbb{E}_{\tau\sim\pib(\cdot\mid x)}[\cdot]\), with analogous notation for
\(\pit\).  We write \(\cumratio_t(\theta)\) when its parameter dependence is
relevant.

\Cref{app:derivation} establishes that \(\cumratio_t\) is the exact
state--action change-of-measure ratio at position \(t\).  Exactness of the
ratio alone, however, does not specify an unbiased gradient estimator: the
advantage and score-function form must also be compatible.  We derive two exact
behavior-rollout representations and then relate them to the proximal
approximations used by the practical \PNPO{} objective.

\subsection{Gradient of the Performance-Difference Identity}

Applying \cref{eq:appendix-change-of-measure} to each state--action
expectation in \cref{eq:appendix-performance-difference} gives
\begin{equation}
  J_x(\pit)-J_x(\pib)
  =
  \mathbb{E}_{\tau\sim\pib(\cdot\mid x)}
  \left[
    \sum_{t=1}^{L}\cumratio_t(\theta)A_t^\beta(s_t,a_t)
  \right].
  \label{eq:appendix-pdl-behavior}
\end{equation}
Because
\begin{align}
  \nabla_\theta\cumratio_t(\theta)
  &=
  \cumratio_t(\theta)\nabla_\theta\log\cumratio_t(\theta)
  \nonumber\\
  &=
  \cumratio_t(\theta)
  \sum_{k=1}^{t}\nabla_\theta\log\pit(a_k\mid s_k)
  \nonumber\\
  &=
  \cumratio_t(\theta)\sum_{k=1}^{t}z_k,
  \label{eq:appendix-ratio-gradient}
\end{align}
and both \(\pib\) and \(A_t^\beta\) are independent of \(\theta\),
differentiating \cref{eq:appendix-pdl-behavior} yields
\begin{equation}
  \nabla_\theta J_x(\pit)
  =
  \mathbb{E}_{\tau\sim\pib(\cdot\mid x)}
  \left[
    \sum_{t=1}^{L}
    \cumratio_t A_t^\beta(s_t,a_t)
    \sum_{k=1}^{t}z_k
  \right].
  \label{eq:appendix-pdl-gradient}
\end{equation}

\subsection{Direct Change of Measure of the Policy Gradient}

The policy-gradient theorem gives
\begin{align}
  \nabla_\theta J_x(\pit)
  &=
  \sum_{t=1}^{L}
  \mathbb{E}_{\substack{s_t\sim d_t^\theta(\cdot\mid x)\\
                        a_t\sim\pit(\cdot\mid s_t)}}
  \left[A_t^\theta(s_t,a_t)z_t\right]
  \nonumber\\
  &=
  \sum_{t=1}^{L}
  \mathbb{E}_{\substack{s_t\sim d_t^\beta(\cdot\mid x)\\
                        a_t\sim\pib(\cdot\mid s_t)}}
  \left[\cumratio_t A_t^\theta(s_t,a_t)z_t\right],
  \label{eq:appendix-direct-gradient}
\end{align}
The second equality applies \cref{eq:appendix-change-of-measure}: the cumulative
ratio changes the sampled state--action distribution from \(\pib\) to \(\pit\),
while \(A_t^\theta z_t\) remains the learner-policy score-function integrand.
We next verify that this current-token-score form and the cumulative-score form
in \cref{eq:appendix-pdl-gradient} agree in expectation.

Reversing the order of summation in \cref{eq:appendix-pdl-gradient} gives
\begin{align}
  &\mathbb{E}_{\pib}
  \left[
    \sum_{t=1}^{L}\cumratio_t A_t^\beta
    \sum_{k=1}^{t}z_k
  \right]
  \nonumber\\
  &\quad =
  \sum_{k=1}^{L}
  \mathbb{E}_{\pib}
  \left[
    z_k\sum_{t=k}^{L}\cumratio_t A_t^\beta
  \right].
  \label{eq:appendix-reordered-gradient}
\end{align}
Let \(\mathcal{F}_k=\sigma(x,a_{1:k})\) denote the prefix information through
and including \(a_k\).  Because \(s_k=(x,a_{<k})\), conditioning on
\(\mathcal{F}_k\) is equivalent to conditioning on \((s_k,a_k)\).  For
\(t>k\),
\begin{equation}
  \cumratio_t
  =
  \cumratio_k\prod_{j=k+1}^{t}\ratio_j.
  \label{eq:appendix-ratio-split}
\end{equation}
It follows that
\begin{align}
  &\mathbb{E}_{\pib}
  \left[
    \left.\cumratio_t A_t^\beta(s_t,a_t)\right|\mathcal{F}_k
  \right]
  \nonumber\\
  &\quad =
  \cumratio_k
  \mathbb{E}_{\pib}
  \left[
    \left.
    \left(\prod_{j=k+1}^{t}\ratio_j\right)
    A_t^\beta(s_t,a_t)
    \right|\mathcal{F}_k
  \right],
  \label{eq:appendix-extract-prefix-ratio}
\end{align}
where \(\cumratio_k\) is \(\mathcal{F}_k\)-measurable.  The remaining
conditional expectation is
\begin{align}
  &\mathbb{E}_{\pib}
  \left[
    \left.
    \left(\prod_{j=k+1}^{t}\ratio_j\right)
    A_t^\beta(s_t,a_t)
    \right|\mathcal{F}_k
  \right]
  \nonumber\\
  &\quad =
  \sum_{a_{k+1:t}}
  \left[
    \prod_{j=k+1}^{t}
    \pib(a_j\mid s_j)
    \frac{\pit(a_j\mid s_j)}{\pib(a_j\mid s_j)}
  \right]
  A_t^\beta(s_t,a_t)
  \nonumber\\
  &\quad =
  \sum_{a_{k+1:t}}
  \left[\prod_{j=k+1}^{t}\pit(a_j\mid s_j)\right]
  A_t^\beta(s_t,a_t)
  \nonumber\\
  &\quad =
  \mathbb{E}_{\pit}
  \left[\left.A_t^\beta(s_t,a_t)\right|\mathcal{F}_k\right].
  \label{eq:appendix-suffix-change}
\end{align}
Substituting \cref{eq:appendix-suffix-change} into
\cref{eq:appendix-extract-prefix-ratio}, for every \(t>k\),
\begin{equation}
  \mathbb{E}_{\pib}
  \left[
    \left.\cumratio_t A_t^\beta(s_t,a_t)\right|\mathcal{F}_k
  \right]
  =
  \cumratio_k
  \mathbb{E}_{\pit}
  \left[\left.A_t^\beta(s_t,a_t)\right|\mathcal{F}_k\right].
  \label{eq:appendix-future-term-change}
\end{equation}
The current term \(\cumratio_k A_k^\beta(s_k,a_k)\) is already determined by
\(\mathcal{F}_k\).  Separating it from the future terms and applying
\cref{eq:appendix-future-term-change} gives
\begin{align}
  &\mathbb{E}_{\pib}
  \left[
    \left.
    \sum_{t=k}^{L}\cumratio_t A_t^\beta(s_t,a_t)
    \right|\mathcal{F}_k
  \right]
  \nonumber\\
  &\quad =
  \cumratio_k A_k^\beta(s_k,a_k)
  +
  \sum_{t=k+1}^{L}
  \mathbb{E}_{\pib}
  \left[
    \left.\cumratio_t A_t^\beta(s_t,a_t)\right|\mathcal{F}_k
  \right]
  \nonumber\\
  &\quad =
  \cumratio_k A_k^\beta(s_k,a_k)
  +
  \cumratio_k\sum_{t=k+1}^{L}
  \mathbb{E}_{\pit}
  \left[\left.A_t^\beta(s_t,a_t)\right|\mathcal{F}_k\right]
  \nonumber\\
  &\quad =
  \cumratio_k
  \mathbb{E}_{\pit}
  \left[
    \left.
    A_k^\beta(s_k,a_k)
    +\sum_{t=k+1}^{L}A_t^\beta(s_t,a_t)
    \right|\mathcal{F}_k
  \right]
  \nonumber\\
  &\quad =
  \cumratio_k
  \mathbb{E}_{\pit}
  \left[
    \left.\sum_{t=k}^{L}A_t^\beta(s_t,a_t)\right|s_k,a_k
  \right].
  \label{eq:appendix-conditional-advantage-sum}
\end{align}

Let \(r_t(s_t,a_t)\) denote the conditional mean immediate reward and set
\(V_{L+1}^\beta=0\).  Under the deterministic token-append transition,
\begin{equation}
  A_t^\beta(s_t,a_t)
  =
  r_t(s_t,a_t)
  +V_{t+1}^\beta(s_{t+1})
  -V_t^\beta(s_t).
  \label{eq:appendix-bellman-advantage}
\end{equation}
Substitution into the last expectation in
\cref{eq:appendix-conditional-advantage-sum} gives
\begin{align}
  &\mathbb{E}_{\pit}
  \left[
    \left.\sum_{t=k}^{L}A_t^\beta(s_t,a_t)\right|s_k,a_k
  \right]
  \nonumber\\
  &\quad =
  \mathbb{E}_{\pit}
  \left[
    \left.
    \sum_{t=k}^{L}
    \bigl(r_t+V_{t+1}^\beta-V_t^\beta\bigr)
    \right|s_k,a_k
  \right]
  \nonumber\\
  &\quad =
  \mathbb{E}_{\pit}
  \left[\left.\sum_{t=k}^{L}r_t\right|s_k,a_k\right]
  -V_k^\beta(s_k)
  \nonumber\\
  &\quad =
  Q_k^\theta(s_k,a_k)-V_k^\beta(s_k),
  \label{eq:appendix-bellman-telescope}
\end{align}
where
\[
  Q_k^\theta(s_k,a_k)
  =
  \mathbb{E}_{\pit}
  \left[\left.\sum_{t=k}^{L}r_t\right|s_k,a_k\right].
\]
Combining
\cref{eq:appendix-conditional-advantage-sum,eq:appendix-bellman-telescope}
yields
\begin{equation}
  \mathbb{E}_{\pib}
  \left[
    \left.\sum_{t=k}^{L}\cumratio_t A_t^\beta\right|\mathcal{F}_k
  \right]
  =
  \cumratio_k
  \left[Q_k^\theta(s_k,a_k)-V_k^\beta(s_k)\right].
  \label{eq:appendix-conditional-q}
\end{equation}
Because \(z_k\) is \(\mathcal{F}_k\)-measurable, the tower property applied to
\cref{eq:appendix-reordered-gradient,eq:appendix-conditional-q} gives
\begin{align}
  &\sum_{k=1}^{L}
  \mathbb{E}_{\pib}
  \left[z_k\sum_{t=k}^{L}\cumratio_t A_t^\beta\right]
  \nonumber\\
  &\quad =
  \sum_{k=1}^{L}
  \mathbb{E}_{\pib}
  \left[
    z_k
    \mathbb{E}_{\pib}
    \left[
      \left.\sum_{t=k}^{L}\cumratio_t A_t^\beta\right|\mathcal{F}_k
    \right]
  \right]
  \nonumber\\
  &\quad =
  \sum_{k=1}^{L}
  \mathbb{E}_{\pib}
  \left[
    \cumratio_k
    \bigl(Q_k^\theta(s_k,a_k)-V_k^\beta(s_k)\bigr)z_k
  \right].
  \label{eq:appendix-tower-gradient}
\end{align}
Now decompose
\begin{equation}
  Q_k^\theta(s_k,a_k)-V_k^\beta(s_k)
  =
  A_k^\theta(s_k,a_k)
  +V_k^\theta(s_k)-V_k^\beta(s_k).
  \label{eq:appendix-q-decomposition}
\end{equation}
The second term depends only on \(s_k\).  Using the state--action factorization
in \cref{eq:appendix-state-action-ratio}, its score contribution vanishes:
\begin{align}
  &\mathbb{E}_{\pib}
  \left[
    \cumratio_k
    \bigl(V_k^\theta(s_k)-V_k^\beta(s_k)\bigr)z_k
  \right]
  \nonumber\\
  &\quad =
  \mathbb{E}_{s_k\sim d_k^\beta(\cdot\mid x)}
  \Bigg[
    \frac{d_k^\theta(s_k\mid x)}{d_k^\beta(s_k\mid x)}
    \bigl(V_k^\theta(s_k)-V_k^\beta(s_k)\bigr)
  \nonumber\\
  &\hspace{22mm}\cdot
    \sum_a\pib(a\mid s_k)
    \frac{\pit(a\mid s_k)}{\pib(a\mid s_k)}
    \nabla_\theta\log\pit(a\mid s_k)
  \Bigg]
  \nonumber\\
  &\quad =
  \mathbb{E}_{s_k\sim d_k^\beta(\cdot\mid x)}
  \left[
    \frac{d_k^\theta(s_k\mid x)}{d_k^\beta(s_k\mid x)}
    \bigl(V_k^\theta(s_k)-V_k^\beta(s_k)\bigr)
    \nabla_\theta\sum_a\pit(a\mid s_k)
  \right]
  \nonumber\\
  &\quad =0.
  \label{eq:appendix-baseline-cancellation}
\end{align}
Substituting
\cref{eq:appendix-q-decomposition,eq:appendix-baseline-cancellation} into
\cref{eq:appendix-tower-gradient} gives
\begin{align}
  &\mathbb{E}_{\pib}
  \left[
    \sum_{t=1}^{L}\cumratio_t A_t^\beta
    \sum_{k=1}^{t}z_k
  \right]
  \nonumber\\
  &\quad =
  \sum_{k=1}^{L}
  \mathbb{E}_{\pib}
  \left[\cumratio_k A_k^\theta(s_k,a_k)z_k\right]
  \nonumber\\
  &\quad =
  \nabla_\theta J_x(\pit).
  \label{eq:appendix-gradient-equivalence}
\end{align}
Hence, \cref{eq:appendix-pdl-gradient,eq:appendix-direct-gradient} are two
exact representations of the same target-policy gradient.  The first pairs
\(A_t^\beta\) with the cumulative prefix score, whereas the second pairs
\(A_t^\theta\) with the current-token score.  Their expectations agree by
\cref{eq:appendix-gradient-equivalence}, but these components are not
interchangeable: in general, \(\cumratio_t A_t^\beta z_t\) is not an exact
target-policy gradient term.
Under the assumptions above and with exact advantages, Monte Carlo estimates
of either representation are unbiased for \(\nabla_\theta J_x(\pit)\).

The corresponding stop-gradient surrogate forms are
\begin{align}
  &\mathbb{E}_{\pib}
  \left[
    \sum_{t=1}^{L}\sg\left[\cumratio_t A_t^\beta\right]
    \sum_{k=1}^{t}\log\pit(a_k\mid s_k)
  \right],
  \nonumber\\[2mm]
  &\mathbb{E}_{\pib}
  \left[
    \sum_{t=1}^{L}\sg\left[\cumratio_t A_t^\theta\right]
    \log\pit(a_t\mid s_t)
  \right].
  \label{eq:appendix-gradient-matching-surrogates}
\end{align}
At the current parameter value, their gradients recover
\cref{eq:appendix-pdl-gradient,eq:appendix-direct-gradient}, respectively.

\subsection{Practical PNPO Surrogate}

The distinction above also clarifies the approximation inherited from the
TRPO/PPO lineage.  In the local surrogate of
\cref{eq:appendix-local-surrogate}, the behavior-policy occupancy and
\(A_t^\beta\) are held fixed, and differentiating the local ratio yields a
current-token score.  Because this surrogate contains \(\ratio_t\) rather than
\(\cumratio_t\), it does not differentiate the full prefix ratio.  TRPO
constrains this local update with a trust region, whereas PPO and \GRPO{} use
clipped local ratios.  \PNPO{} changes the weighting statistic but retains the
corresponding proximal advantage approximation in its current-token-score form:
\begin{equation}
  A_t^\theta(s_t,a_t)
  \approx
  A_t^\beta(s_t,a_t).
  \label{eq:appendix-advantage-approximation}
\end{equation}
In the reported objective, the rollout-derived group-relative signal
\(\adv_{i,t}\) serves as this behavior-side proxy.  As established above,
pairing this proxy with the current-token score is a proximal approximation
rather than the exact \(A_t^\beta\) cumulative-score representation.  \PNPO{}
further replaces the exact cumulative ratio with the prefix-normalized policy
weight
\begin{equation}
  \pnweight_{i,t}
  =
  \cumratio_{i,t}^{1/t}.
  \label{eq:appendix-pn-weight}
\end{equation}
For \(t>1\), replacing \(\cumratio_{i,t}\) with
\(\cumratio_{i,t}^{1/t}\) does not preserve either exact gradient identity;
the normalization deliberately trades exact change of measure for control of
the cumulative log-weight scale.  Using the response-level notation and gate
\(M_{i,t}\) defined in \cref{sec:pnpo-objective}, the reported surrogate is
\begin{align}
  \mathcal{J}_{\PNPO}(\theta)
  =
  \mathbb{E}_{\substack{x\sim\mathcal{D}\\
  y_1,\ldots,y_G\overset{\mathrm{i.i.d.}}{\sim}\pib(\cdot\mid x)}}
  \Bigg[
    \frac{1}{G}\sum_{i=1}^{G}\frac{1}{L_i}
    \sum_{t=1}^{L_i}
    \sg\left[M_{i,t}\pnweight_{i,t}\adv_{i,t}\right]
    \log\pit(y_{i,t}\mid x,y_{i,<t})
  \Bigg].
  \label{eq:appendix-pnpo-surrogate}
\end{align}
Thus, the practical \PNPO{} update combines the current-token score with a
behavior-side advantage proxy, the prefix-normalized weight, position-dependent
gating, and response-level aggregation.  Together, these choices make the
complete objective a deliberately biased proximal surrogate of the
target-policy gradient, while \(\cumratio_t\) remains its exact
change-of-measure reference.

\clearpage
\section{Full Experimental Configuration}
\label{app:configuration}

\begin{table}[h]
  \centering
  \caption{Configuration used for the reported experiments.}
  \label{tab:full-configuration}
  \footnotesize
  \setlength{\tabcolsep}{4pt}
  \begin{tabular}{@{}p{0.31\linewidth}p{0.63\linewidth}@{}}
    \toprule
    Item & Configuration \\
    \midrule
    Base model & DeepSeek-R1-Distill-Qwen-1.5B \\
    Training data & DAPO-Math-17k \\
    Rollout batch & \(256\) prompts \(\times\,8\) responses per prompt \\
    PPO minibatch & \(64\) prompt groups (\(512\) response sequences) \\
    Policy-update epochs & \(1\) or \(4\) \\
    Maximum lengths & \(1{,}024\) prompt tokens; \(15{,}360\) response tokens \\
    Advantage estimator & Group-relative outcome advantage (as in \GRPO{}) \\
    Actor optimization & Learning rate \(10^{-6}\); 10-step warmup; weight
    decay \(0.1\) \\
    Rollout sampling & Temperature \(1.0\) \\
    Evaluation sets & AMC 2023, AIME 2024, AIME 2025 \\
    Evaluation sampling & \(32\) responses per problem; temperature \(0.7\);
    top-\(p\) sampling with \(p=0.9\) \\
    Evaluation schedule & Every \(50\) steps (one-epoch setting); every \(10\)
    steps (four-epoch setting) \\
    Reported metrics & Avg@32 for each benchmark; macro Avg@32 as the
    unweighted mean across benchmarks \\
    Runs & One run for each combination of method and epoch setting (seed
    \(42\)) \\
    \midrule
    \GRPO{} & Token-mean aggregation; local-ratio clip tolerances
    \((0.2,0.28)\); dual-clip coefficient \(10.0\) \\
    \GSPO{} & Mean of token-level score terms within each response, then mean
    across responses; sequence-ratio clip tolerances
    \((3\times10^{-4},4\times10^{-4})\) \\
    \PNPO{} & Mean of token-level score terms within each response, then mean
    across responses; position-dependent gate with base tolerances
    \((7\times10^{-4},9.5\times10^{-4})\) \\
    \bottomrule
  \end{tabular}
\end{table}

\FloatBarrier

\end{document}